\documentclass[11pt,a4paper]{article}
\usepackage[hyperref]{acl2021}
\usepackage{times}
\usepackage{latexsym}
\usepackage{bm}
\usepackage{color,colortbl}
\usepackage{stfloats}
\usepackage{subfigure}
\usepackage{todonotes}
\usepackage{enumitem}
\usepackage{booktabs}
\usepackage{nccmath}
\usepackage{multirow}
\usepackage{subfigure}
\usepackage{makecell}

\usepackage{microtype}

\aclfinalcopy 

\definecolor{LightGray}{gray}{0.85}
\usepackage{soul}

\title{Thank you BART! \\ Rewarding Pre-Trained Models Improves Formality Style Transfer}

\author{
Huiyuan Lai, Antonio Toral, Malvina Nissim\\
CLCG, University of Groningen / The Netherlands\\
\texttt{\{h.lai, a.toral.ruiz, m.nissim\}@rug.nl}
}

\date{}

\begin{document}
\maketitle


\begin{abstract}
Scarcity of parallel data causes 
formality style transfer models to have scarce success in preserving content. We show that fine-tuning pre-trained language (GPT-2) and sequence-to-sequence (BART) models boosts content preservation, and that this is possible even with limited amounts of parallel data.
Augmenting these models with rewards that target style and content --the two core aspects of the task-- we achieve a new state-of-the-art.
\end{abstract}


\section{Introduction and Background}\label{s:intro}
Style transfer is the task of automatically converting a text of one style into another, such as turning the formal ``\emph{I viewed it and I believe it is a quality program.}'' into the  informal ``\emph{I've watched it and it is AWESOME!!!!}''.  This task, which can be used for, e.g., personalised response generation, translation of ancient text into modern text, and text simplification, is particularly challenging since style must be changed while ensuring that content is preserved. 
Accordingly, the performance of  style transfer systems is commonly assessed on both  style strength and content preservation. 

Due to the general scarcity of parallel data, unsupervised approaches are popular. These include disentangling style and content by learning a distinct representation for each \citep{Shen2017, Fu2018StyleTI, john-2019}, and back translation \citep{Zhirui-2018, lample2019multipleattribute, fuli-2019, prabhumoye-2018}. A common strategy to enhance style accuracy is to introduce a reward in the form of a style classifier \cite{lample2019multipleattribute, gong-2019, fuli-2019, wu-2019, Abhilasha-2020}.  
As a result, unsupervised models achieve good accuracy in style strength. Content preservation is 
however usually unsuccessful \citep{rao-tetreault-2018}.

Parallel data can help to preserve content, but is limited. \citet{niu-multi} combine the train sets of two different domains and incorporate machine translation to train their models with a multi-task learning schema, plus model ensembles. \citet{Abhilasha-2020} use it to train a supervised sequence-to-sequence model, and in addition to the commonly used style strength reward, they include a reward based on BLEU~\cite{papineni-2002} to enhance content preservation. \citet{shang-2019} propose a semi-supervised model combining parallel data with large amounts of non-parallel data.



Pre-trained models, successful in a variety of  NLP  tasks, have recently been used in formality style transfer. \citet{zhang-parallel} propose several data augmentation methods for pre-training a transformer-based \citep{Ashish-2017} model and then used gold data for fine-tuning.  Using GPT-2 \citep{radford-2019}, \citet{wang-harn} and \citet{wang-form} propose a  harness-rule-based preprocessing method, and joint training of bi-directional transfer and auto-encoding with two auxiliary losses. Contemporary work by \citet{chawla--semi} develops a  semi-supervised model based on BART large \citep{lewis-etal-2020-bart}.

\paragraph{Contributions}
Focusing specifically on \textit{formality transfer}, for which parallel data is available, (i) we take the contribution of pre-trained models a step further by augmenting them with reward strategies that target content and style, thereby achieving new state-of-the-art results.
(ii) We analyse separately the contribution of pre-trained models on content and style, showing that they take care of preserving content  (the hardest part of style transfer to date), while ensuring style strength.
(iii) Moreover, experimenting with  training size, we show that while parallel data contributes to content preservation, 
fine-tuning pre-trained models with 10\% of parallel data is more successful than training on 100\% of data from scratch. Reducing the need for parallel data opens up the applicability of supervised style transfer to new scenarios: tasks, domains, languages.\footnote{All code at \url{https://github.com/laihuiyuan/Pre-trained-formality-transfer}.}

\section{Method}
\label{sec:method}
We propose a framework 
to control the style of output text for style transfer atop pre-trained models. Given a source sentence $\bm{x}=\{x_{1}, \cdots, x_{n}\}$ of length $n$ with style $s_{1}$ and a target style sentence $\bm{y}=\{y_{1}, \cdots, y_{m}\}$ of length $m$ with style $s_{2}$, our 
model aims to learn two conditional distributions, altering the style of a sentence while preserving its original content. Our framework consists of (i) fine-tuning pre-trained models on a formality  transfer parallel corpus; (ii) incorporating  rewards 
to enhance style change and content preservation.

\begin{figure*}[t]
    \centering
    \subfigure[Architecture of the GPT-2-based model]{ 
       \begin{minipage}{7.5cm}
       \centering 
       \includegraphics[scale=0.7]{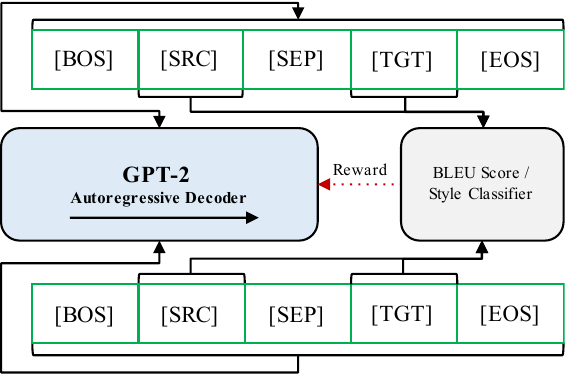}
       \end{minipage}
       \label{fig:gpt2-overview}
    }
    \subfigure[Architecture of the BART-based model]{
    \begin{minipage}{7.5cm}
    \centering 
    \includegraphics[scale=0.7]{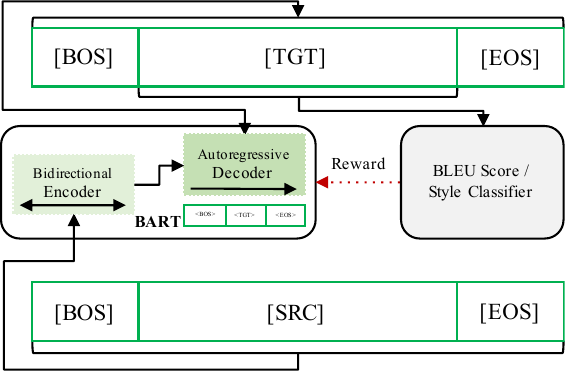} 
    \end{minipage}
    \label{fig:bart-overview}
    }
    \caption{Model architectures. We use three special symbols: [BOS] in front of every source sentence, [SEP] between the source and target sentences (only  in GPT-2), and [EOS] at the end of every target sentence.}
    \label{fig:model-overview}
\end{figure*}

\subsection{Models}

\paragraph{GPT-2}
\label{sub:pre}
This model~\citep{radford-2019} is a transformer-based network \citep{Ashish-2017}.
Given a sentence of tokens $\bm{x}=\{x_{1}, \cdots, x_{l}\}$, the standard language modeling objective is to minimize the following negative log likelihood:
\begin{equation}
\label{loss-gpt}
    L_(\phi)=
    -\Sigma_{i}\textrm{log}(p(x_{i}|x_{i-k:i-1};\phi))
\end{equation}
where $k$ is the size of the context window.
 
To make GPT-2 rephrase a text in the target style, the input pair  	$\left \langle \textrm{Source Sentence, Target Sentence}\right \rangle$ is represented as a
single sequence with three special tokens to mark beginning [BOS] and end [EOS] of every sequence, and to separate source and target sentences [SEP] (Fig.~\ref{fig:gpt2-overview}).
During inference, we feed to GPT-2 the source sentence with [BOS] and [SEP] to infer the target sentence.

\paragraph{BART}
 This is a denoising autoencoder for pretraining sequence-to-sequence models \citep{lewis-etal-2020-bart}. Given a source sentence $\bm{x}$ and a target sentence $\bm{y}$, the loss function is the cross-entropy between the decoder’s output and the target sentence:
\begin{equation}
\label{loss-bart}
    L_(\phi)=
    -\Sigma_{i}\textrm{log}(p(y_{i}|y_{1:i-1}, \bm{x};\phi))
\end{equation}

\subsection{Rewards}
\label{sub:rewards}
Atop the models, we implement two rewards, used in isolation and together, 
to enhance style strength (Style Classification Reward) and content preservation (BLEU Score Reward).

\paragraph{Style Classification Reward}
\label{sub:cls-reward}
As often done in previous work (see Section~\ref{s:intro}), we use a classification confidence reward to encourage larger change in the confidence of a style classifier (SC). 
We pre-train the binary style classifier TextCNN \citep{kim-2014} and use it to evaluate how well the transferred sentence $\bm{y}'$ matches the target style. SC's confidence is formulated as
\begin{equation}
\label{cls-softmax}
    p(s_{i}|\bm{y}') = softmax_{i}(TextCNN(\bm{y}', \theta))
\end{equation}
where i = \{1,2\}, and represent source and target style respectively. $\theta$ are the parameters of the style classifier, fixed during fine-tuning. The reward is
\begin{equation}
\label{reward-cls}
    R_{cls}=\lambda_{cls}[p(s_{2}|\bm{y}') - p(s_{1}|\bm{y}')]
\end{equation}

\noindent where $\bm{y}'$ is the generated target sentence sampled from the model's distribution at each time step in decoding. For the GPT-2 based model, we also add a classification confidence reward to the source sentence, similar to Eq.~\ref{reward-cls}, 
since the model generates sentence $\bm{x}'$ with the original style while generating the target sentence:

\begin{equation}
\label{reward-cls0}
    R_{{cls}_{source}}=\lambda_{cls}[p(s_{1}|\bm{x}') - p(s_{2}|\bm{x}')]
\end{equation}

\paragraph{BLEU Score Reward}
Following \citet{Abhilasha-2020}, we introduce a BLEU-based reward to foster content preservation as in Eq.~\ref{reward-bleu}, where $\bm{y}'$ is the target style text obtained by greedily maximizing the distribution of model outputs at each time step, and $\bm{y}^{s}$ is sampled from the distribution.
\begin{equation}
\label{reward-bleu}
\begin{small}
    R_{bleu}=
    \lambda_{bleu}[bleu(\bm{y}', \bm{y})-bleu(\bm{y}^{s}, \bm{y})]
    \end{small}
\end{equation}

\paragraph{Gradients and Objectives}
The rewards are used for policy learning. The policy gradient\footnote{Additional details are provided in the Appendix.} is
\begin{equation}
\label{gradient-policy}
    \nabla_{\phi}J(\phi)=
    E[R\cdot\nabla_{\phi}log(P(\bm{y}^{s}|\bm{x};\phi))]
\end{equation}
where $R$ is the SC reward and/or the BLEU reward, 
$\bm{y}^{s}$ is sampled from the distribution of model outputs at each decoding time step, and $\phi$ are the parameters of the model. Similarly, we add the policy gradient regarding the source sentence for the SC reward (only for the GPT-2-based model). 

The overall objectives for $\phi$ are the loss of the base model (Eq.~\ref{loss-gpt} or Eq.~\ref{loss-bart}) and the policy gradient of 
the different rewards (Eq.~\ref{gradient-policy}). 

\section{Experiments}

\paragraph{Dataset}

Grammarly’s Yahoo Answers Formality Corpus (GYAFC) \citep{rao-tetreault-2018} is a formality style transfer dataset with parallel formal and informal sentences from two domains: Entertainment \& Music (E\&M) and Family \& Relationships (F\&R). Table~\ref{citation-GYAFC} shows the number of sentences in train, validation, and test. Four human references exist for every valid/test sentence.

\begin{table}[t]
\centering
\small
{
\begin{tabular}{cc|cc|cc}
\toprule
 \multicolumn{2}{c|}{} & \multicolumn{2}{c|}{0 $\longrightarrow$ 1} & \multicolumn{2}{c}{1 $\longrightarrow$ 0}\\
\midrule
  Domain & Train  & Valid  & Test  & Valid  & Test \\
\midrule
  F\&R   & 51,967 & 2,788 & 1,332 & 2,247 & 1,019\\
  E\&M   & 52,595 & 2,877 & 1,416 & 2,356 & 1,082\\
\bottomrule
\end{tabular}}
\caption{\label{citation-GYAFC}
GYAFC dataset. 0~=~informal; 1~=~formal.
}
\end{table}

\paragraph{Setup}

All 
 experiments are implemented 
atop Huggingface's transformers \citep{wolf-etal-2020-transformers}.
Our base models are the GPT-2-based model (117M parameters) and BART-based model (base with 139M parameters and large with 406M). 
We fine-tune them with the Adam optimiser \citep{diederik-kingma-2015}
 with batch size 32; the initial learning rates are $5e^{-5}$ (GPT-2) and $3e^{-5}$ (BART). The final values for $\lambda$ are set to 1 for SC and 0.2 for BLEU based on validation results.
We use early stopping (patience 3) if validation  performance does not improve. Test results are reported with the best validation settings. 

\begin{figure}[!]
    \centering
    \subfigure[GPT-2-based (E\&M)]{ 
       \begin{minipage}{3.5cm}
       \includegraphics[scale=0.23]{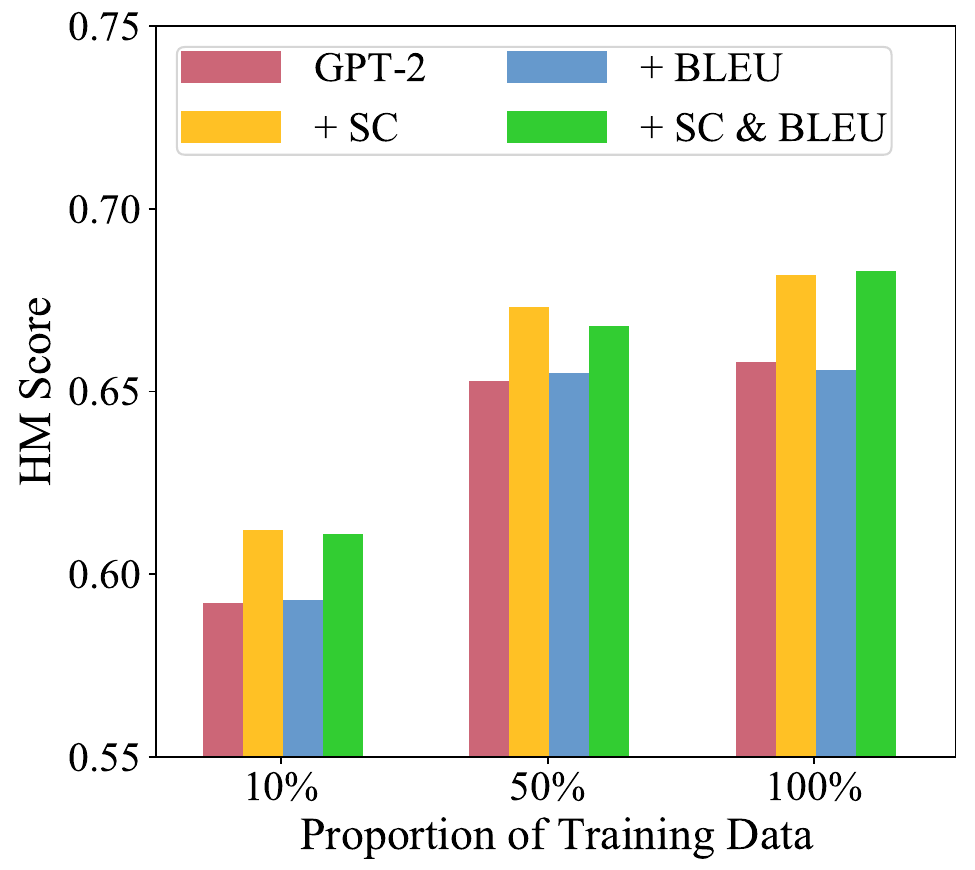}
       \end{minipage}
       \label{fig:eval-GPT2}
    }
    \subfigure[BART-based (E\&M)]{
    \begin{minipage}{3.5cm}
    \includegraphics[scale=0.23]{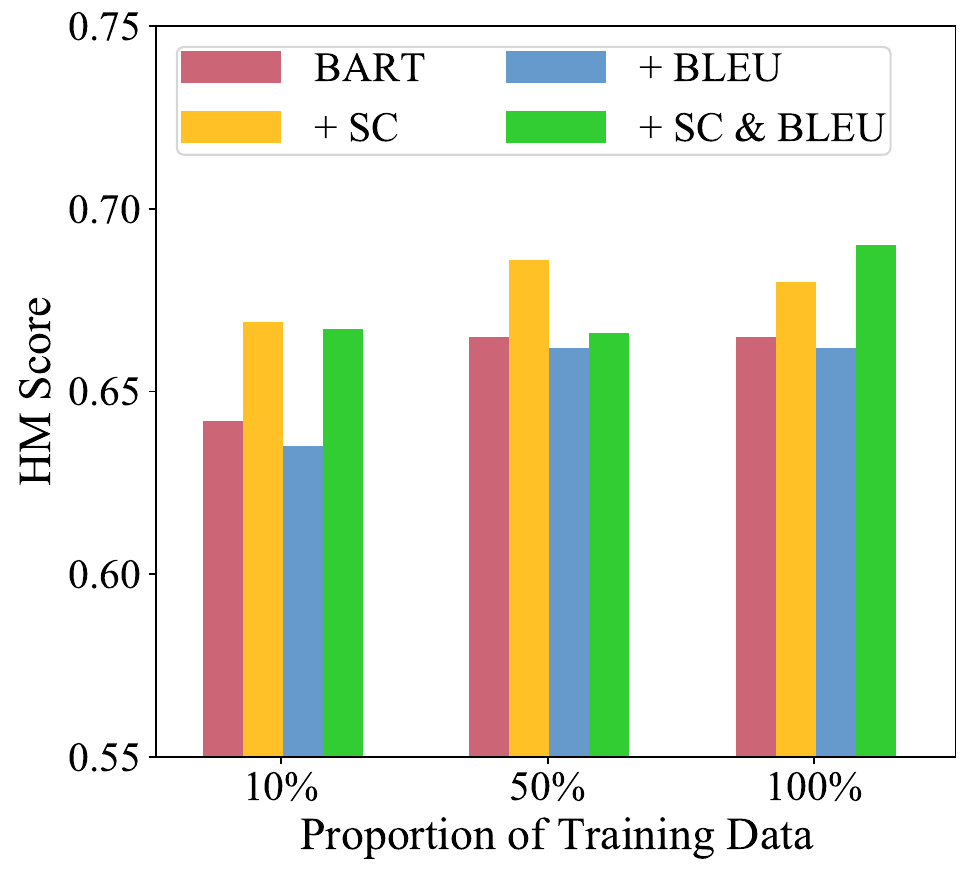} 
    \end{minipage}
    \label{fig:eval-BART}
    }
    \subfigure[GPT-2-based (F\&R)]{ 
       \begin{minipage}{3.5cm}
       \includegraphics[scale=0.23]{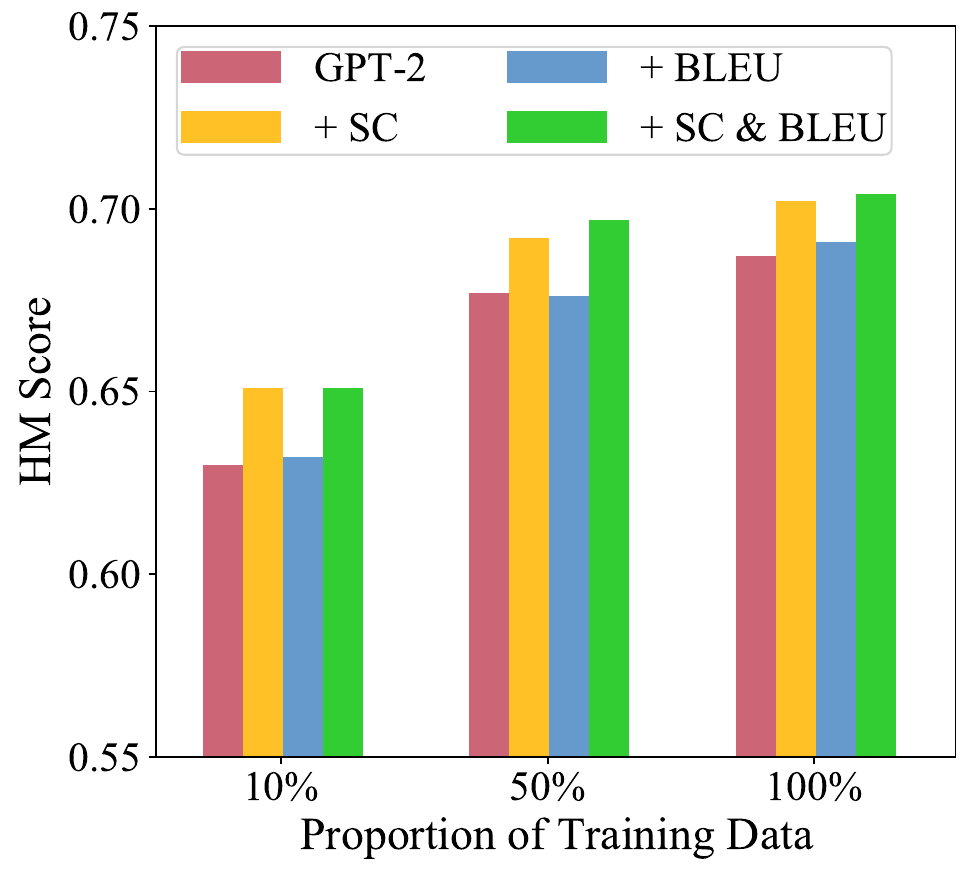}
       \end{minipage}
       \label{fig:eval-GPT2}
    }
    \subfigure[BART-based (F\&R)]{
    \begin{minipage}{3.5cm}
    \includegraphics[scale=0.23]{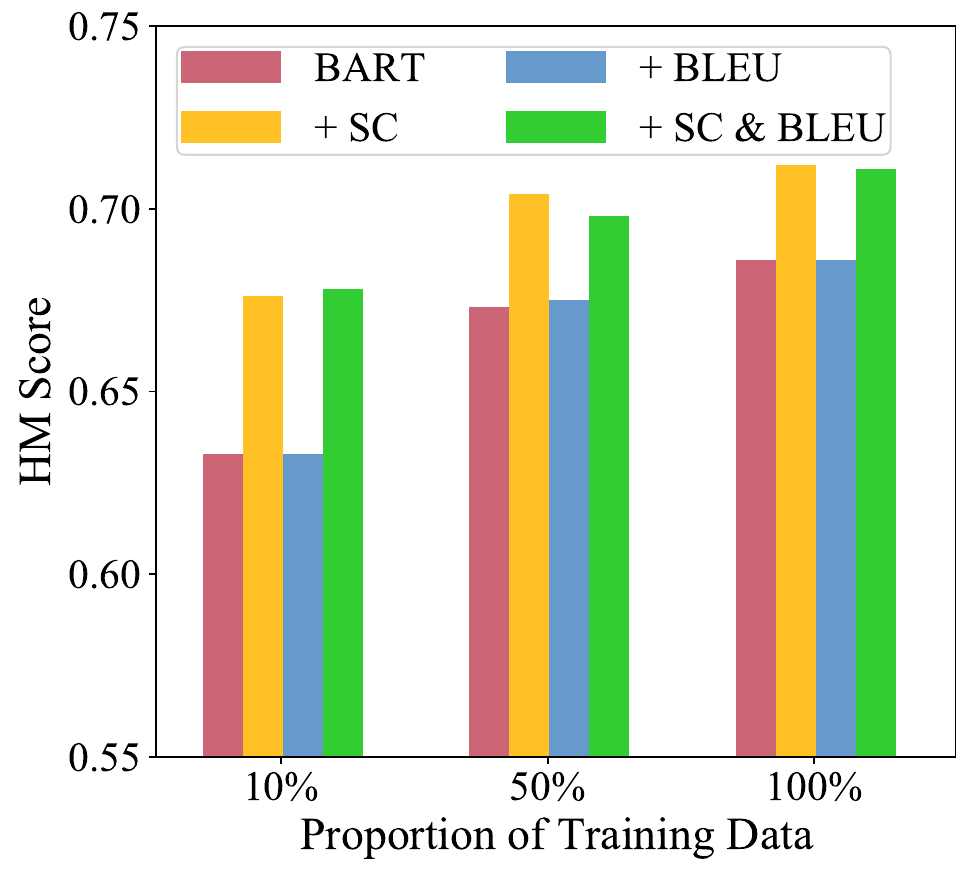} 
    \end{minipage}
    \label{fig:eval-BART}
    }
    \caption{HM score of x\%-sized training sets of GPT-2-/BART-based models with different rewards (none, +SC, +BLEU, +SC \& BLEU) for the two domains (E\&M and F\&R).}
    \label{fig:evaluations}
\end{figure}

\paragraph{Evaluation}
Following previous work \cite{fuli-2019, he-2020, Abhilasha-2020}, we adopt the following strategies. The binary classifier TextCNN \citep{kim-2014} is pre-trained to evaluate style strength; on the human references it has an accuracy of 87.0\% (E\&M) and 89.3\% (F\&R). Based on the four human references, we calculate BLEU\footnote{We use \texttt{multi-bleu.perl} with default settings.} for content preservation. As overall score we compute the harmonic mean (HM) of style accuracy and BLEU. For our evaluation we also test BLEURT, a recent metric for content preservation which correlates better with human judgments than other metrics that take semantic information into account, e.g. METEOR \citep{sellam-etal-2020-bleurt}.

\paragraph{Baselines}
We train a basic supervised model (a Bi-LSTM with attention from  OpenNMT \citep{opennmt}), to assess the impact of the size of parallel training data. We compare our models to the 
five baselines from \citet{rao-tetreault-2018}, and to the best performing formality style transfer methods that report results on the datasets we use. These are mentioned in Section~\ref{s:intro} and summarised as follows:
Bi-directional FT~\citep{niu-multi}, CPLS~\citep{shang-2019}, GPT-CAT~\citep{wang-harn}, S2S-SLS (GPT-2)~\citep{wang-form}, Transformer (data augmentation)~\citep{zhang-parallel}, TS$\rightarrow$CP~\citep{Abhilasha-2020}, and Chawla's \citep{chawla--semi}.
Since supervised methods significantly outperform unsupervised approaches, results for the latter are not considered as the baseline in our experiment..
Disentanglement-based methods are not included since \citet{lample2019multipleattribute} provide evidence that they are surpassed.


\begin{table*}[ht]
\centering
\resizebox{\linewidth}{!}{%
\begin{tabular}{c|l|cccc|@{\hspace{1em}}|l|cccc}
\toprule[2pt]
 Domain & Model & BLEURT & BLEU & ACC & HM & Model & BLEURT & BLEU & ACC & HM \\
\toprule
\multirow{10}{*}{E\&M} & OpenNMT + SC \& BLEU (10\% data)   & -0.919 & 0.231 & 0.886 & 0.366 & OpenNMT + SC \& BLEU (100\% data)     & -0.420 & 0.403 & 0.804 & 0.537\\
\cline{2-11}
\rowcolor{LightGray}
\cellcolor{white}
   & \multicolumn{5}{c|@{\hspace{1em}}|}{\textbf{\textsc{(a) informal $\leftrightarrow$ formal}}} & \multicolumn{5}{c}{\textbf{\textsc{(b) informal $\longrightarrow$ formal}}}\\
\cline{2-11} 
 & NMT-Combined \cite{rao-tetreault-2018}         & -0.100 & 0.501 & 0.797 & 0.615 & GPT-CAT (train on E\&M and F\&R, \citet{wang-harn})       & 0.176 & 0.725 & 0.876 & 0.793\\
 & GPT-2 + SC \& BLEU (10\% data, \textbf{Ours})  & -0.058 & 0.495 & 0.799 & 0.611 & Chawla's (\citet{chawla--semi})                           & 0.260 & 0.762 & 0.910 & 0.829\\
 & GPT-2 + SC \& BLEU (100\% data, \textbf{Ours}) & -0.007 & 0.542 & \textbf{0.923} & 0.683 & BART + SC \& BLEU (train on E\&M, \textbf{Ours})   & 0.218 & 0.730 & 0.887 & 0.801\\
 & BART + SC \& BLEU(10\% data, \textbf{Ours})    & -0.030 & 0.547 & 0.855 & 0.667 & BART + SC \& BLEU (train on E\&M and F\&R, \textbf{Ours})  & 0.236 & 0.745 &\textbf{0.937}&0.830\\
 & BART + SC \& BLEU (100\% data, \textbf{Ours})  & \textbf{0.044} & \textbf{0.577} & 0.859 & \textbf{0.690} & BART large + SC \& BLEU (train on E\&M and F\&R, \textbf{Ours}) & \textbf{0.274} & \textbf{0.765} & 0.929 & \textbf{0.839}\\
\cline{2-11} 
\rowcolor{LightGray}
\cellcolor{white}
 & \multicolumn{5}{c|@{\hspace{1em}}|}{\textbf{\textsc{(c) informal $\leftrightarrow$ formal \& combined domains}}} & \multicolumn{5}{c}{\textbf{\textsc{(d) BLEU evaluated against the first reference}}}\\
\cline{2-11} 
 & Bi-directional FT \cite{niu-multi}                  & 0.023  & 0.554 & 0.818 & 0.661 & *TS$\rightarrow$CP (\citet{Abhilasha-2020})                & - & 0.292 & - & -\\
 & BART large + SC \& BLEU (100\% data, \textbf{Ours}) & \textbf{0.078} & \textbf{0.596} & \textbf{0.905} & \textbf{0.719} & BART + SC \& BLEU (100\% data, \textbf{Ours}) & - & \textbf{0.306} & - & -\\
\toprule[2pt]
\multirow{10}{*}{F\&R} & OpenNMT + SC \& BLEU (10\% data)    & -0.706   & 0.303 & 0.859 & 0.448 & OpenNMT + SC \& BLEU (100\% data)     & -0.304 & 0.477 & 0.789 & 0.595\\
\cline{2-11} 
\rowcolor{LightGray}
\cellcolor{white}
 & \multicolumn{5}{c|@{\hspace{1em}}|}{\textbf{\textsc{(a) informal $\leftrightarrow$ formal}}}  & \multicolumn{5}{c}{\textbf{\textsc{(b) informal $\longrightarrow$ formal}}}\\
\cline{2-11} 
 & NMT-Combined \cite{rao-tetreault-2018}         & -0.089 & 0.527 & 0.798 & 0.635 & *GPT-CAT (train on E\&M and F\&R, \citet{wang-harn} )         & -     & 0.769 & - & -\\   
 & GPT-2 + SC \& BLEU (10\% data, \textbf{Ours})  & -0.027 & 0.528 & 0.849 & 0.651 & Chawla's (\citet{chawla--semi}) & 0.302 & \textbf{0.799} & 0.910 & 0.851\\
 & GPT-2 + SC \& BLEU (100\% data, \textbf{Ours}) & 0.038  & 0.572 & \textbf{0.915} & 0.704 & BART + SC \& BLEU (train on F\&R, \textbf{Ours})  & 0.271 & 0.770 & 0.897 & 0.829\\
 & BART + SC \& BLEU (10\% data, \textbf{Ours})   & 0.039  & 0.571 & 0.833 & 0.678 & BART + SC \&  BLEU (train on F\&R and E\&M, \textbf{Ours}) & 0.270 & 0.777 & 0.912 & 0.839\\
 & BART + SC \& BLEU (100\% data, \textbf{Ours})  & \textbf{0.068}  & \textbf{0.595}& 0.882 & \textbf{0.711} & BART large + SC \&  BLEU (train on F\&R and E\&M, \textbf{Ours}) & \textbf{0.324} & 0.793 & \textbf{0.920} & \textbf{0.852}\\
\cline{2-11} 
\rowcolor{LightGray}
\cellcolor{white}
 & \multicolumn{5}{c|@{\hspace{1em}}|}{\textbf{\textsc{(c) informal $\leftrightarrow$ formal \& combined domains}}} & \multicolumn{5}{c}{\textbf{\textsc{(d) 10\% parallel training data}}}\\
\cline{2-11} 
 & Bi-directional FT (\citet{niu-multi}  & 0.037 & 0.568 & 0.839 & 0.677 & *CPLS \citep{shang-2019}                                  & - & 0.379 & - & -\\
 & BART large + SC \& BLEU (100\% data, \textbf{Ours)} & \textbf{0.100} & \textbf{0.611} & \textbf{0.900} & \textbf{0.728} & BART + SC \& BLEU \textbf{(Ours)}& - & \textbf{0.571} & -\\
\bottomrule[2pt]
\end{tabular}}
\caption{\label{citation-results}
Comparison of our models to previous work. The best score for each metric in each block is boldfaced. Notes:  
(i) if the output of previous work is available, we re-calculate the scores using our evaluation metrics. Otherwise, scores are from the paper and we mark this with (*); (ii) (B) shows our results on informal-to-formal to compare with \citet{wang-harn} and \citet{chawla--semi}, who only transfer in this direction; (iii) in (C) we train on the concatenated data from both domains, to compare against~\citet{niu-multi}; (iv) in (E\&M (D)) we re-evaluate our system against the first reference only, as done by \citet{Abhilasha-2020}.}
\end{table*}


\paragraph{Results} 
Figure~\ref{fig:evaluations} shows the HM score of x\%-sized training sets on the E\&M and the F\&R domains. 
Increasing train set size from 10\% to 50\% has a greater boost on GPT-2-based models than BART's.
However, BART-based models obtain the highest results.
Table~\ref{citation-results} reports a selection of our models \footnote{In the table we report results for the models that use both rewards (BLEU and SC) since this setting mostly leads to best results. Complete results for all models (and sample outputs) are in the Appendix.} and previous state-of-the-art work.
Zooming in on the single measures, we see in Table~\ref{citation-results} how
varying training size 
reveals the impact of parallel data on content preservation: OpenNMT's BLEU score on E\&M increases from 
0.231 with 10\% of the data to 0.403 with 100\%. 
Style accuracy appears instead easier to achieve even with limited supervision. Increasing training size for fine-tuning either pre-trained model  does not however yield dramatic improvements in content preservation (e.g. from 0.547 to 0.577 BLEU for BART base on E\&M). In fact, fine-tuning a pre-trained model (either GPT-2 or BART) with just 10\% of parallel data, leads to better content preservation (0.547 BLEU with BART on E\&M) than OpenNMT with 100\% (0.403). This suggests that content preservation is largely taken care of by the pre-trained models, already, and  
%
%
can explain why the BLEU-based reward does not help too much in isolation (see Fig.~\ref{fig:evaluations}).
Conversely, the SC reward consistently boosts style accuracy in both BART and GPT-2. Nevertheless, combining rewards can be beneficial. 
Overall, BART-based models perform better on content preservation while results on style strength are mixed.

Given the experimental setup of some previous work, we ran additional comparisons (blocks (B), (C), and (D) of Table~\ref{citation-results}).
In all cases, our results are higher than the previous state-of-the-art. For example, in F\&R (D) our model with 10\% parallel data outperforms \citet{shang-2019}'s semi-supervised model, which uses about 9.5\% parallel data and large amounts of non-parallel data (BLEU 0.571 vs 0.379). Fine-tuning BART on both domains (C)\footnote{Following~\citet{Catherine-2017}, we add a token to each training instance that specifies its domain.} leads to the best results to date on both datasets (E\&M: 0.719; F\&R: 0.728).

With respect to the two evaluation metrics used for content preservation (BLEU and BLEURT),
we can observe in Table~\ref{citation-results} that they follow a similar trend.
In fact, they correlate very highly (Pearson's $r=.951$, p$<$.001, $n=14$ for E\&M, and $r=.951$, p$<$.001, $n=13$ for F\&R).

\begin{table*}[!ht]
\centering
\tiny
\resizebox{\linewidth}{!}{%
\begin{tabular}{l|l|ccc}
\toprule[2pt]
System & Sentence & BLEURT & BLEU & ACC\\
\hline
\rowcolor{LightGray}
\multicolumn{5}{c}{\textbf{\textsc{From informal to formal}}}\\
\hline
 Source              & i say omarion.he has the hair clothes and body,a triple deal on one person. & \multicolumn{3}{c}{-}\\
  Reference 1         & My choice is Omarion as he has high quality, hair, clothes, and body to create a triple deal in one person. & \multicolumn{3}{c}{-}\\
 Reference 2         & I would say Omarion because he has the hair, clothes, and body; A triple deal on a single person. & \multicolumn{3}{c}{-}\\
 Reference 3         & I pick Omarion, he has the hair, the clothes, and the body.  A triple deal on one person. & \multicolumn{3}{c}{-}\\
 Reference 4         & Omarion has the hair, clothes, and the body. & \multicolumn{3}{c}{-}\\
 \hline
 PBMT-Combined \citep{rao-tetreault-2018}                 & \textcolor{red}{I say omarion}. \textcolor{red}{he} has the hair, clothes and body, the deal on one person.            & -0.153 & 0.509 & 0.946\\
 Bi-directional FT \citep{niu-multi}                    & \textcolor{red}{I say} Omarion, he has the hair clothes and body, and a triple deal on one person.   & -0.149 & 0.510 & 0.953\\
 GPT-CAT \citep{wang-harn}     & \textcolor{red}{I say} Omarion. He has the hair, clothes, and body, a triple deal on one person.     & 0.044  & 0.585 & \textbf{1.000}\\
 S2S-SLS \citep{wang-form}                       & \textcolor{red}{I say} Omarion. He has the hair clothes and body, a triple deal on one person.       & -0.035 & 0.350 & \textbf{1.000}\\
 Transformer \citep{zhang-parallel} & \textcolor{red}{I say omarionhe} has the hair clothes and body, a triple deal on one person.         & -0.255 & 0.462 & 0.892\\
 Chawla's \citep{chawla--semi}                           & \textcolor{red}{I say Marion} because he has the hair, clothes and body, a triple deal on one person.& -0.538 & 0.534 & 0.989\\
\hline
OpenNMT + SC \& BLEU (\textbf{Ours}) & \textcolor{red}{I say} Omarion. He has the hair clothes and body.                     & -0.325 & 0.147 & \textbf{1.000}\\
GPT-2 + SC \& BLEU (\textbf{Ours})      & \textcolor{red}{I say} Omarion. He has the hair clothes and body, a triple deal on one person.  & -0.035 & 0.350 & \textbf{1.000}\\
BART base + SC \& BLEU (\textbf{Ours})  & I would say \textcolor{red}{Omar}. He has the hair, clothes, and body. It is a triple deal on one person. & -0.012 & 0.589 & \textbf{1.000}\\
BART large + SC \& BLEU (\textbf{Ours}) & I would say Omarion. He has the hair, clothes, and body, a triple deal on one person. & \textbf{0.096} & \textbf{0.657} & \textbf{1.000}\\
\hline
\rowcolor{LightGray}
\multicolumn{5}{c}{\textbf{\textsc{From formal to informal}}}\\
\hline
 Source              & I suggest avoiding hot dogs, and not watching this movie with your little sister. & \multicolumn{3}{c}{-}\\
\hline
 Reference 1         & Don't eat hot dogs, or watch this movie with your little sister! & \multicolumn{3}{c}{-}\\
 Reference 2         & Don't do hot dogs or this movie with your kid sister. & \multicolumn{3}{c}{-}\\
 Reference 3         & don't eat hot dogs and don't watch it w/ ur lil sis! & \multicolumn{3}{c}{-}\\
 Reference 4         & Don't eat hot dogs or watch this flick with your lil sis! & \multicolumn{3}{c}{-}\\
\hline
PBMT-Combined \citep{rao-tetreault-2018}& I suggest avoiding hot dogs, and not watching this movie with your little sister. & -0.298 & 0.417 & 0.004\\
Bi-directional FT \citep{niu-multi}     & I suggest avoiding hot dogs and not watching this movie with your little sister.  & -0.233 & 0.437 & 0.009\\
\hline
OpenNMT with SC \& BLEU             & Can't watch this movie with your little sister.                                       & -0.521 & 0.542 & 0.783\\
GPT-2 + SC \& BLEU                  & don't watch this movie with your little sister.                                       & -0.415 & 0.599 & \textbf{1.000}\\
BART + SC \& BLEU                   & avoid hot dogs and not watch this movie with your little sister.                      & \textbf{-0.016} & 0.610 & 0.925\\
BART large + SC \& BLEU & Avoid hot dogs and don't watch this movie with your little sister.                                & -0.171 & \textbf{0.800} & 0.825\\
\bottomrule[2pt]
\end{tabular}}
\caption{\label{tab:example-eval-level}
Sample model outputs and their sentence-level scores on the E\&M domain, where red denotes improperly generated words or content. Note that ACC indicates style confidence here.
}
\end{table*}

\paragraph{Finer-grained Analysis}
Table~\ref{tab:example-eval-level} shows example outputs and their evaluation according to the metrics we use; the outputs are produced by existing systems we compare to, and our own models.\footnote{More examples are in Appendix.} 

In the ``Informal to Formal" example, we can see that text generated by most systems is assessed with a high confidence in style conversion, except for PBMT-Combined \citep{rao-tetreault-2018} and Transformer \citep{zhang-parallel} (the name ``omarionhe" should be ``Omarion", and the word ``he" at the beginning of the sentence should be ``He"). However, the sentences generated by previous systems are not so fluent, and some of them fail in preserving content (Transformer \citep{zhang-parallel} (``omarionhe") and Chawla's \citep{chawla--semi} (``Marion")).
For our models, the Bi-LSTM based model fails in content preservation while the systems based on pre-trained models are much better at this task. Our model based on BART~Large generates this specific sentence accurately in terms of content preservation, style strength, and fluency. 

When looking at the ``Formal to Informal" example in Table~\ref{tab:example-eval-level},
we observe that the two  previously existing  systems  replace very little (one comma by the Bi-directional FT \citep{niu-multi}) or nothing at all (PBMT-Combined \citep{rao-tetreault-2018}).
Conversely, our systems make substantial modifications, resulting in output sentences that are noticeably more informal than the input sentence. OpenNMT and the GPT-2-based models lose part of the content (the suggestion to avoid hot dogs) while the two BART-based systems manage to preserve the whole message.


\section{Conclusions}

Fine-tuning pre-trained models proves a successful strategy for formality style transfer, especially towards content preservation, thereby reducing the need for parallel data.
A sequence-to-sequence pre-trained model (BART) outperforms a language model (GPT-2) in content preservation, and overall, and with the addition of rewards achieves new state-of-the-art results. 
The fact that GPT-2 is instead often better at style strength could be (partly) due to how the style reward is implemented in the two models (Eq.~\ref{reward-cls} and~\ref{reward-cls0}), and will need further investigation. 
For a better understanding of the different behaviour of BART and GPT-2 for this task, the next natural step is to include human evaluation.

\section*{Acknowledgments}


This work was partly funded by the China Scholarship Council (CSC). The anonymous ACL reviewers provided us with useful comments which contributed to improving this paper and its presentation, so we're grateful to them. We would also like to thank the Center for Information
Technology of the University of Groningen for their
support and for providing access to the Peregrine
high performance computing cluster.

\section*{Impact Statement}
All work that automatically generates and/or alters natural text could unfortunately be used maliciously. While we cannot fully prevent such uses once our models are made public, we do hope that writing about risks explicitly and also raising awareness of this possibility in the general public are ways to contain the effects of potential harmful uses. We are open to any discussion and suggestions to minimise such risks.


\bibliographystyle{acl_natbib}
\bibliography{anthology,acl2021}

\clearpage

\setcounter{page}{1}

\onecolumn

\setcounter{table}{0}
\renewcommand{\thetable}{A.1.\arabic{table}}
\setcounter{figure}{0}
\renewcommand{\thefigure}{A.1.\arabic{figure}}

\appendix
\section{\Large Appendices 
}

\bigskip

This Appendices include: 1) detailed results for all experiments (\ref{append:results}); 2) more details on policy gradient (\ref{append:policy}); 3) some example outputs of various models and their sentence-level scores, to give an idea of what the generated sentences look like when style transfer is applied. We specifically focus on the 100\% parallel data settings for our models (\ref{append:examples}).

\vspace*{0.5cm}


\subsection{Detailed Results of Models}
\label{append:results}

We report here the full set of results for all our models and previous work. \\ 

\subsection*{(a) Detailed Results of Our Models}

\begin{table*}[ht]
\centering
\resizebox{\linewidth}{!}{%
\begin{tabular}{l|cccc|cccc|cccc}
\toprule[2pt]
 Model & BLEURT & BLEU & ACC & HM & BLEURT & BLEU & ACC & HM & BLEURT & BLEU & ACC & HM \\
 \toprule
 Proportion of parallel training data & \multicolumn{4}{c|}{10\%} & \multicolumn{4}{c|}{50\%} & \multicolumn{4}{c}{100\%}\\
 \toprule
OpenNMT (Bi-LSTM)     & -0.919 & 0.231 & 0.886 & 0.366 & -0.489 & 0.392 & 0.789 & 0.524 & \textbf{-0.420} & 0.403 & 0.804 & 0.537\\
OpenNMT + SC          & \textbf{-0.902} & \textbf{0.238} & \textbf{0.893} & \textbf{0.376} & -0.500 & 0.386 & \textbf{0.821} & 0.526 & -0.451 & 0.399 & 0.789 & 0.530\\
OpenNMT + BLEU        & -0.926 & 0.232 & 0.888 & 0.368 & \textbf{-0.485} & 0.389 & 0.800 & 0.523 & -0.485 & \textbf{0.412} & 0.767 & 0.536\\
OpenNMT + SC \& BLEU  & -0.903 & 0.234 & 0.890 & 0.371 & -0.497 & \textbf{0.391} & 0.813 & \textbf{0.528} & -0.442 & 0.403 & \textbf{0.810} & \textbf{0.538}\\
 \hline
GPT-2 base            & -0.042 & 0.492 & 0.741 & 0.592 & 0.004  & \textbf{0.541} & 0.825 & 0.653 & \textbf{0.006}  & \textbf{0.549} & 0.821 & 0.658\\
GPT-2 + SC            & -0.048 & 0.492 & \textbf{0.810} & \textbf{0.612} & -0.014 & 0.531 & \textbf{0.919} & \textbf{0.673} & -0.001 & 0.543 & 0.917 & 0.682\\
GPT-2 + BLEU          & \textbf{-0.041} & \textbf{0.497} & 0.735 & 0.593 & \textbf{0.006}  & 0.539 & 0.833 & 0.655 & 0.005  & 0.546 & 0.822 & 0.656\\
GPT-2 + SC \& BLEU    & -0.058 & 0.495 & 0.799 & 0.611 & -0.014 & 0.530 & 0.903 & 0.668 & -0.007 & 0.542 & \textbf{0.923} & \textbf{0.683}\\ 
 \hline
BART base             & \textbf{0.035}  & \textbf{0.547} & 0.776 & 0.642 & 0.036  & \textbf{0.572} & 0.794 & 0.665 & 0.048  & \textbf{0.578} & 0.784 & 0.665\\
BART + SC             & 0.021  & 0.539 & \textbf{0.882} & \textbf{0.669} & 0.035  & 0.566 & \textbf{0.872} & \textbf{0.686} & 0.045  & 0.571 & 0.841 & 0.680\\
BART + BLEU           & 0.034  & 0.541 & 0.769 & 0.635 & 0.040  & 0.567 & 0.796 & 0.662 & \textbf{0.050}  & 0.576 & 0.777 & 0.662\\
BART + SC \& BLEU     & 0.030  & \textbf{0.547} & 0.855 & 0.667 & \textbf{0.042}  & 0.562 & 0.817 & 0.666 & 0.044  & 0.577 & \textbf{0.859} & \textbf{0.690}\\
 \hline
BART large + SC \& BLEU & 0.035 & 0.560 & 0.847 & 0.674 & 0.070 & 0.585 & 0.900 & 0.709 & 0.072 & 0.584 & 0.886 & 0.704\\

 \toprule
 \multicolumn{13}{c}{\textbf{\textsc{Combined two domains without domain tag}}}\\
 \toprule
BART base             & \textbf{0.038} & \textbf{0.559} & 0.731 & 0.634 & 0.050 & \textbf{0.581} & 0.795 & 0.671 & \textbf{0.054} & \textbf{0.585} & 0.809 & 0.679\\
BART + SC             & 0.031 & 0.546 & \textbf{0.830} & 0.659 & 0.043 & 0.575 & \textbf{0.865} & \textbf{0.691} & 0.039 & \textbf{0.585} & \textbf{0.884} & \textbf{0.704}\\
BART + BLEU           & 0.033 & 0.555 & 0.743 & 0.635 & 0.042 & 0.575 & 0.810 & 0.673 & \textbf{0.054} & 0.583 & 0.814 & 0.679\\
BART + SC \& BLEU     & 0.024 & 0.556 & 0.815 & \textbf{0.661} & \textbf{0.054} & 0.578 & 0.845 & 0.685 & 0.050 & 0.580 & 0.859 & 0.692\\
 \hline
BART large + sc \& BLEU & 0.071 & 0.576 & 0.867 & 0.692 & 0.075 & 0.593 & 0.887 & 0.711 & 0.086 & 0.597 & 0.888 & 0.714\\

 \toprule
 \multicolumn{13}{c}{\textbf{\textsc{Combined two domains with domain tag}}}\\
 \toprule
BART base             & \textbf{0.042} & 0.552 & 0.754 & 0.637 & 0.054 & 0.579 & 0.748 & 0.653 & \textbf{0.060} & 0.582 & 0.787 & 0.669\\
BART + SC             & 0.035 & 0.555 & 0.831 & 0.666 & 0.039 & 0.571 & 0.833 & 0.678 & 0.046 & 0.579 & \textbf{0.895} & \textbf{0.703}\\
BART + BLEU           & 0.039 & 0.554 & 0.745 & 0.635 & \textbf{0.056} & 0.578 & 0.745 & 0.651 & 0.049 & \textbf{0.588} & 0.825 & 0.685\\
BART + SC \& BLEU       & 0.039 & \textbf{0.556} & \textbf{0.845} & \textbf{0.671} & 0.046 & \textbf{0.580} & \textbf{0.834} & \textbf{0.684} & 0.047 & 0.583 & 0.883 & 0.702\\
\hline
BART large + SC \& BLEU & 0.077 & 0.575 & 0.793 & 0.667 & 0.073 & 0.587 & 0.870 & 0.701 & 0.078 & 0.596 & 0.905 & 0.719\\

\bottomrule[2pt]
\end{tabular}}
\caption{\label{citation-emresults}
Evaluation results of x\%-sized training sets (10\%, 50\% and 100\%) on the E\&M domain. The best score for each metric in each table section is boldfaced. BLEURT scores are calculated based on the BLEURT-base model with 128 tokens. Note that 
(i) Both BLEURT and BLEU are calculated against the four human references; (ii) ACC is the accuracy of the output labeled as the target style by the binary classifier; and (iii) HM is the harmonic mean of ACC and BLEU.}
\end{table*}

\clearpage

\begin{table*}[!ht]
\centering
\resizebox{\linewidth}{!}{%
\begin{tabular}{l|cccc|cccc|cccc}
\toprule[2pt]
 Model & BLEURT & BLEU & ACC & HM & BLEURT & BLEU & ACC & HM & BLEURT & BLEU & ACC & HM \\
 \toprule
 Proportion of parallel training data & \multicolumn{4}{c|}{10\%} & \multicolumn{4}{c|}{50\%} & \multicolumn{4}{c}{100\%}\\
 \toprule
OpenNMT (Bi-LSTM)       & -0.706 & 0.303 & 0.859 & 0.448 & -0.304 & 0.449 & 0.792 & 0.573 & -0.304 & 0.477 & 0.789 & 0.595\\
OpenNMT + SC            & \textbf{-0.695} & \textbf{0.322} & \textbf{0.860} & \textbf{0.469} & -0.337 & 0.447 & 0.838 & \textbf{0.583} & -0.289 & 0.466 & 0.824 & 0.595\\
OpenNMT + BLEU          & -0.712 & 0.311 & 0.829 & 0.452 & \textbf{-0.292} & \textbf{0.455} & 0.808 & 0.582 & \textbf{-0.246} & \textbf{0.478} & 0.789 & 0.595\\
OpenNMT + SC \& BLEU    & -0.699 & 0.320 & 0.828 & 0.462 & -0.332 & 0.444 & \textbf{0.847} & \textbf{0.583} & -0.288 & 0.472 & \textbf{0.848} & \textbf{0.606}\\
 \hline
GPT-2 base              & -0.020 & \textbf{0.531} & 0.775 & 0.630 & \textbf{0.027} & \textbf{0.567} & 0.841 & 0.677 & \textbf{0.046} & 0.576 & 0.850 & 0.687\\
GPT-2 + SC              & -0.031 & 0.529 & 0.847 & \textbf{0.651} & 0.020 & 0.563 & 0.897 & 0.692 & 0.031 & 0.569 & \textbf{0.916} & 0.702\\
GPT-2 + BLEU            & \textbf{-0.016} & 0.529 & 0.786 & 0.632 & 0.026 & 0.566 & 0.838 & 0.676 & 0.041 & \textbf{0.577} & 0.860 & 0.691\\
GPT-2 + SC \& BLEU      & -0.027 & 0.528 & \textbf{0.849} & \textbf{0.651} & 0.015 & 0.562 & \textbf{0.917} & \textbf{0.697} & 0.038 & 0.572 & 0.915 & \textbf{0.704}\\
 \hline
BART base               & \textbf{0.045} & 0.565 & 0.719 & 0.633 & 0.071 & 0.589 & 0.786 & 0.673 & \textbf{0.080} & 0.600 & 0.801 & 0.686\\
BART + SC               & 0.041 & 0.569 & \textbf{0.833} & 0.676 & 0.061 & \textbf{0.592} & \textbf{0.869} & \textbf{0.704} & 0.067 & 0.601 & 0.874 & \textbf{0.712}\\
BART + BLEU             & 0.041 & 0.566 & 0.719 & 0.633 & \textbf{0.072} & 0.590 & 0.789 & 0.675 & 0.078 & \textbf{0.602} & 0.798 & 0.686\\
BART + SC \& BLEU       & 0.039 & \textbf{0.571} & \textbf{0.833} & \textbf{0.678} & 0.057 & 0.589 & 0.858 & 0.698 & 0.068 & 0.595 & \textbf{0.882} & 0.711\\
 \hline
BART large + SC \& BLEU & 0.095 & 0.585 & 0.816 & 0.681 & 0.087 & 0.604 & 0.891 & 0.720 & 0.095 & 0.615 & 0.876 & 0.722 \\

 \toprule
 \multicolumn{13}{c}{\textbf{\textsc{Combined two domains without domain tag}}}\\
 \toprule
BART base               & \textbf{0.035} & \textbf{0.572} & 0.734 & 0.643 & 0.060 & 0.592 & 0.821 & 0.688 & \textbf{0.074} & 0.604 & 0.807 & 0.691\\
BART + SC               & 0.026 & 0.563 & \textbf{0.821} & 0.668 & 0.056 & 0.592 & \textbf{0.890} & \textbf{0.711} & 0.054 & 0.602 & \textbf{0.877} & \textbf{0.714}\\
BART + BLEU             & 0.033 & 0.568 & 0.732 & 0.640 & \textbf{0.064} & 0.593 & 0.834 & 0.693 & 0.073 & \textbf{0.606} & 0.831 & 0.701\\
BART + SC \& BLEU       & 0.028 & \textbf{0.57}2 & 0.812 & \textbf{0.671} & 0.054 & \textbf{0.596} & 0.843 & 0.698 & 0.063 & 0.601 & 0.872 & 0.712\\
 \hline
BART large + SC \& BLEU & 0.087 & 0.598 & 0.869 & 0.708 & 0.094 & 0.607 & 0.871 & 0.715 & 0.100 & 0.610 & 0.889 & 0.724\\

 \toprule
  \multicolumn{13}{c}{\textbf{\textsc{Combined two domains with domain tag}}}\\
 \toprule
 BART base              & 0.042 & 0.570 & 0.779 & 0.658 & \textbf{0.072} & 0.592 & 0.768 & 0.669 & \textbf{0.078} & 0.604 & 0.801 & 0.689\\
 BART + SC              & 0.035 & \textbf{0.574} & 0.849 & \textbf{0.685} & 0.058 & 0.586 & \textbf{0.861} & 0.697 & 0.059 & 0.599 & 0.892 & 0.718\\
 BART + BLEU            & \textbf{0.047} & 0.572 & 0.761 & 0.653 & 0.071 & 0.591 & 0.772 & 0.669 & 0.077 & \textbf{0.605} & 0.817 & 0.695\\
 BART + SC \& BLEU      & 0.043 & 0.573 & \textbf{0.850} & \textbf{0.685} & 0.057 & \textbf{0.595} & 0.849 & \textbf{0.700} & 0.064 & 0.603 & \textbf{0.896} & \textbf{0.721}\\
 \hline
 BART large + SC \& BLEU & 0.089 & 0.590 & 0.801 & 0.679 & 0.099 & 0.604 & 0.869 & 0.713 & 0.100 & 0.611 & 0.900 & 0.728\\
 
\bottomrule[2pt]
\end{tabular}}
\caption{\label{citation-frresults}
Evaluation results of x\%-sized training sets (10\%, 50\% and 100\%) on the F\&R domain. The best score for each metric in each table section is boldfaced. BLEURT scores are calculated based on the BLEURT-base model with 128 tokens. Note that 
(i) Both BLEURT and BLEU are calculated against the four human references; (ii) ACC is the accuracy of the output labeled as the target style by the binary classifier; and (iii) HM is the harmonic mean of ACC and BLEU.}
\end{table*}


\clearpage
\subsection*{(b) Comparison of our models with the other models}

\begin{table*}[ht]
\centering
\resizebox{\linewidth}{!}{%
\begin{tabular}{c|l|cccc|@{\hspace{1em}}|l|cccc}
\toprule[2pt]
 Domain & Model & BLEURT & BLEU & ACC & HM & Model & BLEURT & BLEU & ACC & HM \\
\toprule
\multirow{15}{*}{E\&M} & \multicolumn{5}{c|@{\hspace{1em}}|}{\textbf{\textsc{(a) informal $\leftrightarrow$ formal}}} & \multicolumn{5}{c}{\textbf{\textsc{(b) informal $\longrightarrow$ formal}}}\\
\cline{2-11}
 & Rule-based \cite{rao-tetreault-2018}          & -0.221 & 0.420 & 0.704 & 0.526 & GPT-CAT (train on E\&M, \citet{wang-harn})          & 0.170 & 0.713 & 0.905 & 0.801\\
 & NMT-baseline \cite{rao-tetreault-2018}        & -0.267 & 0.437 & 0.851 & 0.577 & GPT-CAT (train on E\&M and F\&R, \citet{wang-harn}) & 0.176 & 0.725 & 0.876 & 0.793\\
 & NMT-copy \cite{rao-tetreault-2018}            & -0.269 & 0.441 & 0.808 & 0.571 & S2S-SLS(\citet{wang-form})                          & 0.173 & 0.711 & 0.919 & 0.802\\
 & NMT-Combined \cite{rao-tetreault-2018}        & -0.100 & 0.501 & 0.797 & 0.615 & Transformer (\citet{zhang-parallel})                & 0.191 & 0.734 & 0.887 & 0.803\\
 & PBMT-Combined \cite{rao-tetreault-2018}       & -0.088 & 0.502 & 0.753 & 0.602 & Chawla's \citep{chawla--semi}                       & 0.260 & 0.762 & 0.910 & 0.829\\
 & GPT-2 + SC \& BLEU (10\% data, \textbf{Ours}) & -0.058 & 0.495 & 0.799 & 0.611 & GPT-2 + SC \& BLEU (train on E\&M, \textbf{Ours})   & 0.159 & 0.701 & 0.927 & 0.798\\
 & GPT-2 + SC \& BLEU (100\% data, \textbf{Ours})& -0.007 & 0.542 & \textbf{0.923} & 0.683 & BART + SC \& BLEU (train on E\&M, \textbf{Ours})   & 0.218 & 0.730 & 0.887 & 0.801\\
 & BART + SC \& BLEU (10\% data, \textbf{Ours})  & 0.030  & 0.547 & 0.855 & 0.667 & BART + SC \& BLEU (train on E\&M and F\&R, \textbf{Ours})  & 0.236 & 0.745 &\textbf{0.937}&0.830\\
 & BART + SC \& BLEU (100\% data, \textbf{Ours}) & \textbf{0.044} & \textbf{0.577} & 0.859 & \textbf{0.690}  & BART large + SC \& BLEU (train on E\&M and F\&R, \textbf{Ours}) & \textbf{0.274} & \textbf{0.765} & 0.929 & \textbf{0.839}\\
 
\cline{2-11}
 & \multicolumn{5}{c|@{\hspace{1em}}|}{\textbf{\textsc{(c) informal $\leftrightarrow$ formal \& combined domains}}} & \multicolumn{5}{c}{\textbf{\textsc{(d) BLEU evaluated against the first reference}}}\\
\cline{2-11}
 & Bi-directional FT \cite{niu-multi}  & 0.023 & 0.554 & 0.818 & 0.661 & *TS$\rightarrow$CP \citep{Abhilasha-2020}               & - & 0.292 & - & -\\
 & BART large + SC \& BLEU (10\% data, \textbf{Ours})         & 0.077 & 0.575 & 0.793 & 0.667 & GPT-2 + SC \& BLEU (100\% data, \textbf{Ours})                                   & - & 0.296 & - & -\\
 & BART large + SC \& BLEU (100\% data, \textbf{Ours})        & \textbf{0.078} & \textbf{0.596} & \textbf{0.905} & \textbf{0.719} & BART + SC \& BLEU (100\% data, \textbf{Ours}) & - & \textbf{0.306} & - & -\\
\toprule[2pt]
\multirow{13}{*}{F\&R} & \multicolumn{5}{c|@{\hspace{1em}}|}{\textbf{\textsc{(a) informal $\leftrightarrow$ formal}}} & \multicolumn{5}{c}{\textbf{\textsc{(b) informal $\longrightarrow$ formal}}}\\
\cline{2-11}
 & Rule-based \cite{rao-tetreault-2018}          & -0.226 & 0.450 & 0.738 & 0.559 & *GPT-CAT (train on F\&R, \citet{wang-harn})           & - & 0.773 & - & -\\
 & NMT-baseline \cite{rao-tetreault-2018}        & -0.183 & 0.500 & 0.818 & 0.621 & *GPT-CAT (train on E\&M and F\&R, \citet{wang-harn})  & - & 0.769 & - & -\\
 & NMT-copy \cite{rao-tetreault-2018}            & -0.186 & 0.492 & 0.807 & 0.611 & S2S-SLS(GPT-2, \citet{wang-form})         & 0.244 & 0.766 & 0.857 & 0.809\\
 & NMT-Combined \cite{rao-tetreault-2018}        & -0.089 & 0.527 & 0.798 & 0.635 & Transformer (\citet{zhang-parallel})      & 0.246 & 0.770 & 0.890 & 0.827\\
 & PBMT-Combined \cite{rao-tetreault-2018}       & -0.062 & 0.517 & 0.788 & 0.624 & Chawla's \citep{chawla--semi}             & 0.302 & \textbf{0.799} & 0.910 & 0.851\\
 & GPT-2 + SC \& BLEU (10\% data, \textbf{Ours)} & -0.027 & 0.528 & 0.849 & 0.651 & GPT-2 + SC \& BLEU  (train on F\&R, \textbf{Ours}) & 0.226 & 0.747 & 0.921 & 0.825\\
 & GPT-2 + SC \& BLEU (100\% data, \textbf{Ours})& 0.038  & 0.572 & \textbf{0.915}& 0.704 & BART + SC \& BLEU (train on F\&R, \textbf{Ours})  & 0.271 & 0.770 & 0.897 & 0.829\\
 & BART + SC \& BLEU (10\% data, \textbf{Ours})   & 0.039  & 0.571 & 0.833 & 0.678 &  BART + SC \& BLEU (train on F\&R and E\&M, \textbf{Ours})& 0.270 & 0.777 & 0.912 & 0.839\\
 & BART + SC \& BLEU (100\% data, \textbf{Ours})  & \textbf{0.068}  & \textbf{0.595}& 0.882 & \textbf{0.711}  & BART large + SC \& BLEU (train on F\&R and E\&M, \textbf{Ours}) & \textbf{0.324} & 0.793 & \textbf{0.920} & \textbf{0.852}\\
\cline{2-11}
 & \multicolumn{5}{c|@{\hspace{1em}}|}{\textbf{\textsc{(c) informal $\leftrightarrow$ formal \& combined domains}}} & \multicolumn{5}{c}{\textbf{\textsc{(d) 10\% parallel training data (from paper)}}}\\
\cline{2-11}
 & Bi-directional FT (\citet{niu-multi}& 0.037 & 0.568 & 0.839 & 0.677 & *CPLS \citep{shang-2019}                                      & - & 0.379 & - & -\\
 & BART large + SC \& BLEU (10\% data, \textbf{Ours})  & 0.089 & 0.590 & 0.801 & 0.679    & GPT-2 + SC \& BLEU  (\textbf{Ours})        & - & 0.528 & - & -\\
 & BART large + SC \& BLEU (100\% data, \textbf{Ours})  & \textbf{0.100} & \textbf{0.611} & \textbf{0.900} & \textbf{0.728} & BART + SC \& BLEU (\textbf{Ours})                              & - & \textbf{0.571} & - & -\\
\bottomrule[2pt]
\end{tabular}}
\caption{\label{citation-emcomparison}
Comparison of our models with the other models. The best score for each metric in each block is boldfaced. 
BLEURT scores are calculated based on the BLEURT-base model with 128 tokens.
Notes:  
(i) if the output of a previous work is available, we re-calculate the scores using our evaluation metrics. Otherwise we take the scores from the paper and mark this with a (*); (ii) in (B) we report our results on informal-to-formal alone to compare with several systems which only transfer in this direction; (iii) in (C) we train systems on the concatenated data from both domains, to compare against~\citet{niu-multi}; (iv) in (E\&M (D)) we re-evaluate our system against the first reference only, as this is what \citet{Abhilasha-2020} do.}
\end{table*}

\subsection{Policy Gradient}
\label{append:policy}
Reinforcement learning (RL) is a sub-field of machine learning that is concerned with how intelligent agents ought to take actions in an environment in order to maximize the cumulative reward. Here, we employ the policy gradient algorithm~\citep{Williams-1992} to maximize the expected reward (style strength and/or content preservation) of the generated sequence $\bm{y}^{s}$, whose gradient with respect to the parameters $\phi$ of the neural network model is estimated by sampling as:

\begin{equation} 
\begin{split}
\nabla_{\phi}J(\phi)&= R\cdot\nabla_{\phi} \sum_{i}  P(\bm{y}_i^{s}|\bm{x}_i;\phi)\\
&= \sum_{i}P(\bm{y}_i^{s}|\bm{x}_i;\phi) R_i \nabla_{\theta} \log(P(\bm{y}_i^{s}|\bm{x}_i;\phi))\\
&\simeq \frac{1}{N}\sum_{i=1}^N R_i \nabla_{\phi} \log(P(\bm{y}_i^{s}|\bm{x}_i;\phi))\\
&= E[R\cdot\nabla_{\phi}log(P(\bm{y}^{s}|\bm{x};\phi))]
\end{split}
\end{equation}



where $J(\cdot)$ is the objective function, $\nabla_{\phi}J(\cdot)$ is the gradient of $J(\cdot)$ with respect to $\phi$, $R_{i}$ is the reward of the $i_{th}$ sequence $\bm{y}^{s}$ that is sampled from the distribution of model outputs at each decoding time step, $\phi$ are the parameters of the model, $N$ is the sample size, and $E(\cdot)$ is the expectation.

Regarding the reward of style classification for GPT-2 based model, we design two rewards (Eq.~\ref{reward-cls} and Eq.~\ref{reward-cls0}) for source sentence and target sentence, respectively. The policy gradient is then
\begin{equation} 
\begin{split}
\nabla_{\phi}J(\phi)&= E[R_{cls_{source}}\cdot\nabla_{\phi}log(P(\bm{y}^{s}_{source}|\bm{x}_{source};\phi))]\\
&+E[R_{cls_{target}}\cdot\nabla_{\phi}log(P(\bm{y}^{s}_{target}|\bm{x}_{source, target};\phi))]
\end{split}
\end{equation}


\clearpage
\setcounter{table}{0}
\renewcommand{\thetable}{A.3.\arabic{table}}

\subsection{Example Outputs of Various Models}
\label{append:examples}

\begin{table*}[!ht]
\centering
\tiny
\resizebox{\linewidth}{!}{%
\begin{tabular}{l|l|ccc}
\toprule
System & \makecell[c]{From informal to formal} & BLEURT & BLEU & ACC\\
\hline
 Source              & i say omarion.he has the hair clothes and body,a triple deal on one person. & \multicolumn{3}{c}{-}\\
\hline
 Reference 1         & My choice is Omarion as he has high quality, hair, clothes, and body to create a triple deal in one person. & \multicolumn{3}{c}{-}\\
 Reference 2         & I would say Omarion because he has the hair, clothes, and body; A triple deal on a single person. & \multicolumn{3}{c}{-}\\
 Reference 3         & I pick Omarion, he has the hair, the clothes, and the body.  A triple deal on one person. & \multicolumn{3}{c}{-}\\
 Reference 4         & Omarion has the hair, clothes, and the body. & \multicolumn{3}{c}{-}\\
\hline
PBMT-Combined \citep{rao-tetreault-2018}& \textcolor{red}{I say} omarion. \textcolor{red}{he} has the hair, clothes and body, the deal on one person.            & -0.153 & 0.509 & 0.946\\
Bi-directional FT \citep{niu-multi}     & \textcolor{red}{I say} Omarion, he has the hair clothes and body, and a triple deal on one person.   & -0.149 & 0.510 & 0.953\\
GPT-CAT \citep{wang-harn}               & \textcolor{red}{I say} Omarion. He has the hair, clothes, and body, a triple deal on one person.     & 0.044  & 0.585 & \textbf{1.000}\\
S2S-SLS \citep{wang-form}               & \textcolor{red}{I say} Omarion. He has the hair clothes and body, a triple deal on one person.       & -0.035 & 0.350 & \textbf{1.000}\\
Transformer \citep{zhang-parallel}      & \textcolor{red}{I say omarionhe} has the hair clothes and body, a triple deal on one person.         & -0.255 & 0.462 & 0.892\\
Chawla's \citep{chawla--semi}           & \textcolor{red}{I say Marion} because he has the hair, clothes and body, a triple deal on one person.& -0.538 & 0.534 & 0.989\\
\hline
OpenNMT                             & He has the hair clothes and body.                                                     & -0.540 & 0.139 & 0.998\\
OpenNMT with SC                     & \textcolor{red}{I say} Omarion, he has the hair clothes and body.                     & -0.389 & 0.558 & 0.969\\
OpenNMT with BLEU                   & \textcolor{red}{I say} Omarion. He has the hair clothes and body.                     & -0.325 & 0.147 & \textbf{1.000}\\
OpenNMT with SC \& BLEU             & \textcolor{red}{I say} Omarion. He has the hair clothes and body.                     & -0.325 & 0.147 & \textbf{1.000}\\
\hline
GPT-2 base                          & \textcolor{red}{I say} Omarion. He has the hair and body, a triple deal on one person.          & -0.087 & 0.342 & \textbf{1.000}\\
GPT-2 + SC                          & \textcolor{red}{I say} Omarion because he has the hair clothes and body.                        & -0.264 & 0.634 & 0.976\\
GPT-2 + BLEU                        & \textcolor{red}{I say} Omarion. He has the hair clothes and body, a triple deal on one person.  & -0.035 & 0.350 & \textbf{1.000}\\
GPT-2 + SC \& BLEU                  & \textcolor{red}{I say} Omarion. He has the hair clothes and body, a triple deal on one person.  & -0.035 & 0.350 & \textbf{1.000}\\
\hline
BART base                           & I would say \textcolor{red}{Omar}. He has the hair, clothes, and body. It is a triple deal on one person. & -0.012 & 0.589 & \textbf{1.000}\\
BART + SC                           & I would say \textcolor{red}{Omar}. He has the hair, clothes, and body. It is a triple deal on one person. & -0.012 & 0.589 & \textbf{1.000}\\
BART + BLEU                         & I would say \textcolor{red}{Omar}. He has the hair, clothes, and body of a triple deal on one person.     & -0.230 & 0.600 & \textbf{1.000}\\
BART + SC \& BLEU                   & I would say \textcolor{red}{Omar}. He has the hair, clothes, and body. It is a triple deal on one person. & -0.012 & 0.589 & \textbf{1.000}\\
\hline
BART large + SC \& BLEU & I would say Omarion. He has the hair, clothes, and body, a triple deal on one person. & \textbf{0.096} & \textbf{0.657} & \textbf{1.000}\\

\toprule[0.8pt]
System & \makecell[c]{From formal to informal} & BLEURT & BLEU & ACC\\
\hline
 Source              & I suggest avoiding hot dogs, and not watching this movie with your little sister. & \multicolumn{3}{c}{-}\\
\hline
 Reference 1         & Don't eat hot dogs, or watch this movie with your little sister! & \multicolumn{3}{c}{-}\\
 Reference 2         & Don't do hot dogs or this movie with your kid sister. & \multicolumn{3}{c}{-}\\
 Reference 3         & don't eat hot dogs and don't watch it w/ ur lil sis!  & \multicolumn{3}{c}{-}\\
 Reference 4         & Don't eat hot dogs or watch this flick with your lil sis! & \multicolumn{3}{c}{-}\\
\hline
PBMT-Combined \citep{rao-tetreault-2018}& I suggest avoiding hot dogs, and not watching this movie with your little sister. & -0.298 & 0.417 & 0.004\\
Bi-directional FT \citep{niu-multi}     & I suggest avoiding hot dogs and not watching this movie with your little sister.  & -0.233 & 0.437 & 0.009\\
\hline
OpenNMT                             & hott dogs and not watching this movie with ur little sister                           & -0.885 & 0.118 & \textbf{1.000}\\
OpenNMT with SC                     & Im not watching this movie with your little sister...I suggest him hot dogs.          & -0.765 & 0.349 & 0.981\\
OpenNMT with BLEU                   & Well, and not watching this movie with your little sister.                            & -0.826 & 0.445 & 0.633\\
OpenNMT with SC \& BLEU             & Can't watch this movie with your little sister.                                       & -0.521 & 0.542 & 0.783\\
\hline
GPT-2 base                          & Don't watch this movie with your little sister.                                       & -0.415 & 0.573 & 0.851\\
GPT-2 + SC                          & don't watch this movie with your little sister.                                       & -0.415 & 0.599 & \textbf{1.000}\\
GPT-2 + BLEU                        & Don't watch this movie with your little sister!                                       & -0.360 & 0.634 & 0.919\\
GPT-2 + SC \& BLEU                  & don't watch this movie with your little sister.                                       & -0.415 & 0.599 & \textbf{1.000}\\
\hline
BART base                           & avoid hot dogs and not watch this movie with your little sister.                      & \textbf{-0.016} & 0.610 & 0.925\\
BART + SC                           & avoid hot dogs and not watch this movie with your little sister.                      & \textbf{-0.016} & 0.610 & 0.925\\
BART + BLEU                         & avoid hot dogs and not watching this movie with your little sister.                   & -0.034 & 0.514 & 0.910\\
BART + SC \& BLEU                   & avoid hot dogs and not watch this movie with your little sister.                      & \textbf{-0.016} & 0.610 & 0.925\\
\hline
BART large + SC \& BLEU & Avoid hot dogs and don't watch this movie with your little sister.                                & -0.171 & \textbf{0.800} & 0.825\\

\bottomrule
\end{tabular}}
\caption{\label{example-eval-all}
Sample model outputs and their sentence-level scores on the E\&M domain, where red denotes improperly generated words or content. Note that ACC indicates style confidence here.
}
\end{table*}

\end{document}